# USING SHADOWS IN CIRCULAR SYNTHETIC APERTURE SONAR IMAGING FOR TARGET ANALYSIS

Yann Le Gall             General Sonar Studies, Thales DMS France SAS, Brest, France
Nicolas Burlet           General Sonar Studies, Thales DMS France SAS, Brest, France
Mathieu Simon            General Sonar Studies, Thales DMS France SAS, Brest, France
Fabien Novella           DGA Naval Techniques, Brest, France
Samantha Dugelay         General Sonar Studies, Thales UK, Templecombe, UK
Jean-Philippe Malkasse General Sonar Studies, Thales DMS France SAS, Brest, France

## 1   INTRODUCTION

Circular Synthetic Aperture Sonar (CSAS) is an innovative adaptation of Synthetic Aperture Sonar (SAS) that leverages circular trajectories around targets to enhance resolution. While conventional SAS systems are limited by the horizontal aperture of the transmitter, CSAS overcomes this limitation by continuously illuminating an area of interest along the entire trajectory. As a result, CSAS achieves a significantly smaller theoretical resolution limit, offering improved imaging capabilities [Pailhas, 2017]. However, achieving this theoretical resolution limit in CSAS can be challenging due to errors in trajectory estimation, resulting in a loss of focus in the synthetic array [Billon and Fohanno, 1998]. The ability to maintain a focused synthetic array throughout the circular trajectory is particularly demanding. To address these challenges, various advancements have been proposed. Autofocusing techniques correct the focus aberration caused by phase error accumulation, resulting in well-focused CSAS images [Callow et al., 2010; Marston et al., 2014; Marston and Kennedy, 2021]. Additionally, multi-look processing [Fortune et al., 2003] has been introduced to reduce image speckle and improve image quality.

CSAS scans produce high-resolution, low-noise images with sharp edges, making them valuable for target identification. This imaging technique finds applications in scenarios where swift reacquisition of critical targets is essential, long standoff distances are required, or poor water clarity precludes the use of optical systems. Despite these advancements, CSAS images tend to flatten information, diluting aspect-dependent responses into a single image. To address this limitation, color-by-aspect (CBA) processing has been proposed, encoding the dominant backscattering angle using hue [Plotnick and Marston, 2018]. However, while this method significantly improves aspect-dependent information, it does not fully recover all aspect-dependent characteristics.

Notably, in CSAS processing, the circular displacement of the illuminator introduces parallax, filling in shadow regions and causing the loss of target shadows on the seafloor. Shadows provide valuable information about target size and shape, which is essential for target recognition. In conventional side-scan sonar, shadow shape analysis plays a crucial role in target recognition processes and automatic recognition algorithms [Fohanno et al., 2010][Myers and Fawcett, 2010]. This paper introduces an interactive viewer that facilitates the exploration of the acoustic response of the target as a function of orientation. Futhermore, we propose a method to retrieve shadow information from CSAS data to enhance target analysis and enable 3D reconstruction. We employ sub-aperture filtering to generate a collection of images from various points of view along the circular trajectory, and we apply shadow enhancement to obtain well-defined shadows. These resulting images are integrated into the interactive viewer, enabling easy probing of target shadows as a function of orientation. Finally, a space-carving reconstruction method is employed to infer the 3D shape of the object from the shadows. The results demonstrate the potential of shadows in circular SAS for improving target analysis and enabling 3D reconstruction.

The outline of this paper is as follows. Section 2 introduces the application of CSAS to SAMDIS array data with conventional multi-look processing and CBA. In Section 3, we introduce our CSAS slider, which allows for interactive exploration of target responses, and our approach for enhanced shadow visualization in CSAS imaging. Finally, in Section 4, we showcase our method for 3D reconstruction using the retrieved shadow information.

Overall, this research aims to advance CSAS imaging techniques by incorporating shadow information, enhancing target analysis, and facilitating accurate 3D reconstruction.

## 2      CSAS APPLICATION TO SAMDIS DATA

The Synthetic Aperture Mine Detection and Imaging System SAMDIS is a versatile mine warfare sonar that offers multiple imaging modes and high-resolution capabilities [Chabah et al., 2016; Delayes et al., 2022; Malkasse et al., 2023]. It can produce real-time imagery up to 2.5cm resolution along the track and 1.25cm resolution across the track. The system addresses the need for confidence in Mine Counter Measure (MCM) operations by employing a multi-aspect single-pass imaging mode that provides performance equivalent to conducting three surveys with different track orientations in a single coverage. This mode enables effective mine hunting in various environments. SAMDIS is adaptable to different platforms and missions, with varying payload lengths that can be integrated onto towed bodies or Autonomous Underwater Vehicles (AUVs).

Additionally, SAMDIS has the potential to support Circular SAS processing. In winter 2019, sea trials were conducted to collect data along a 360° circular trajectory with a 75m radius around a known target. The target is depicted in Figure 1-(a). It is positioned at the center of a $3m \times 3m$ grid which consists of individual $20cm \times 50cm$ concrete blocks. The acquired data underwent Circular SAS processing. The results are presented in Figure *1*-(b)(c). The Figure *1*-(b) displays the Circular SAS image after applying multi-look processing to reduce speckle, while Figure *1*-(c) displays the color-by-aspect Circular SAS image.

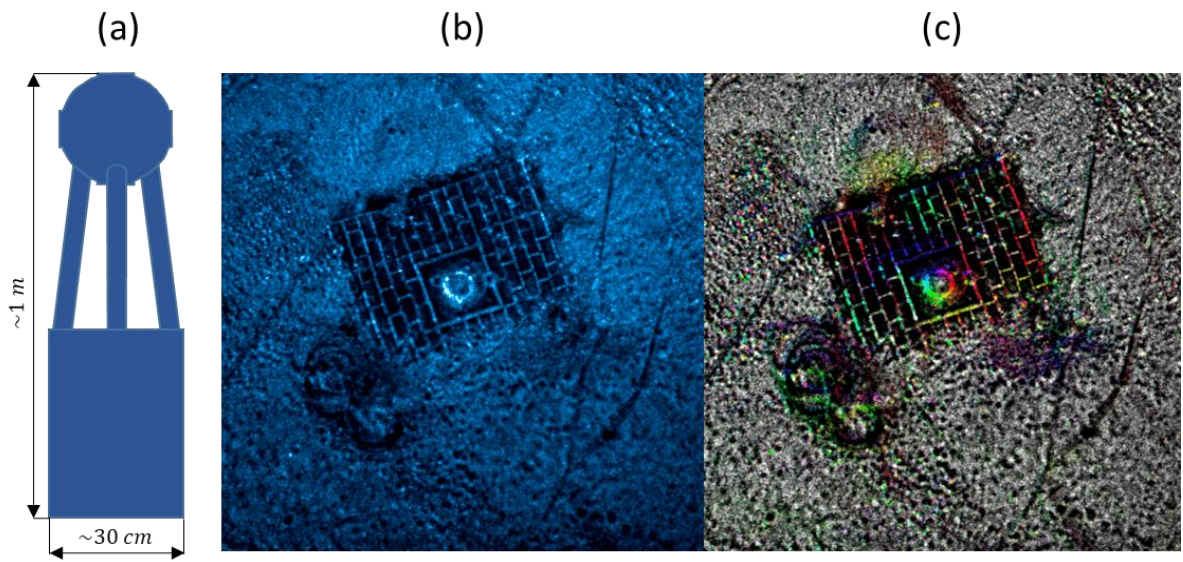

Figure 1: (a) Schematic representation of the target in the center of the sonar image, (b) Multi-look processed CSAS image, and (c) Color-by-aspect CSAS image.

These images offer valuable insights into the characteristics of the target and contribute to a better understanding of the seabed scene. The color-by-aspect image particularly emphasizes anisotropic scattering, and therefore the visualization of topography and specific elements of the scene such as the central targets, the laying grid, and certain cables. However, several crucial scene features lack clarity or fail to be adequately represented. We find that the method does not allow highlighting phenomena that appear over a very short orientation range. For instance, thin cables may exhibit strong responses but only within a limited angular range. Additionally, due to the elimination of projected shadows, it becomes challenging to discern the vertical characteristics of the target and other distinctive elements.

## 3    SHADOW ENHANCED CSAS

### 3.1    CSAS slider

Sub-aperture images can be efficiently obtained from the coherent complex CSAS image using spectral masking [Marston et al., 2011]. To create a sub-aperture, wavenumbers within the range $[\theta_0 - \Delta\theta, \theta_0 + \Delta\theta]$ are retained, and then transformed back into the image domain through a 2D inverse fast Fourier transform. This process essentially simulates observing the scene from a restricted range of angles centered around the angle $\theta_0$. Sub-aperture imaging offers the capability to isolate and analyze individual responses of a scene from a specific viewpoint, offering a valuable approach to investigate the acoustic response of the scene in relation to the orientation. However, visualizing individual sub-aperture images without an appropriate tool can be laborious and may affect analysis accuracy.

To enable ergonomic visualization of the acoustic response of the scene as a function of orientation, we introduce an interactive viewer in the form of a circular slider. The circular slider facilitates the selection of the sub-aperture that is being viewed by moving the cursor along a circle representing the circular trajectory. We have applied sub-aperture filtering and incorporated the resulting images into the circular slider representation for the SAMDIS circular data mentioned in the previous section. In this particular case, a sub-aperture image is generated every degree, with an aperture width of 12 degrees. The choice of the aperture width is determined to achieve a high image resolution while avoiding dilution of brief angular responses. Spectral domain sub-aperture filtering at angle $\theta_0$ and resulting sub-aperture image within the CSAS slider are illustrated in Figure 2. The green cursor of the CSAS slider depicted in Figure 2-(b) enables to switch smoothly between viewing angles.

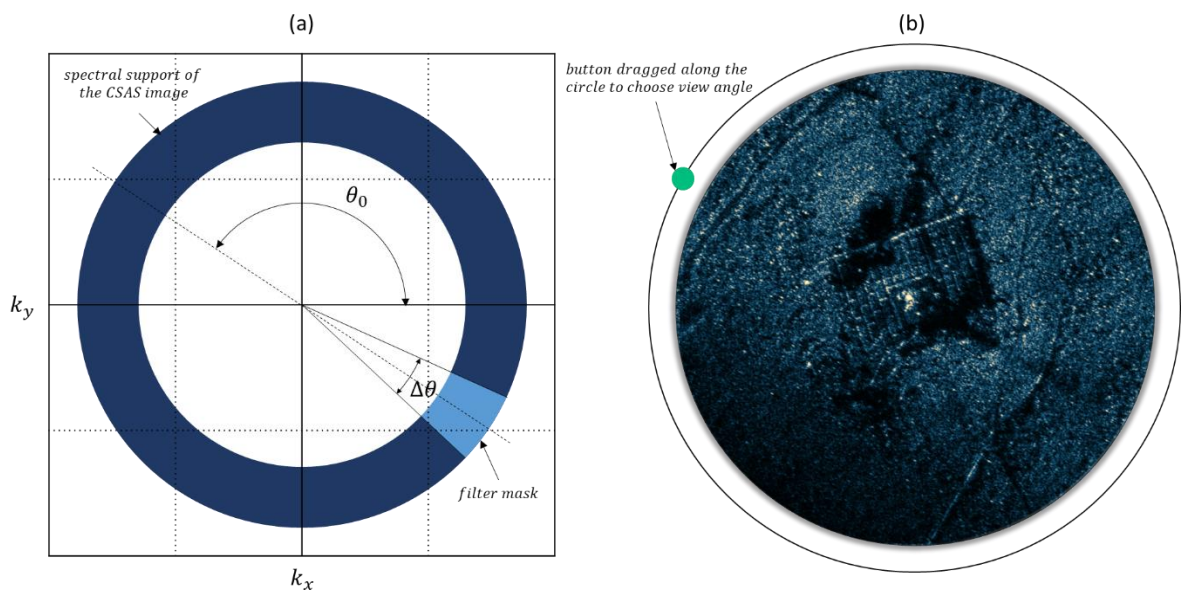

Figure 2: (a) Illustration of sub-aperture filtering at angle $\theta_0$ by masking of the 2D Fourier transform, and (b) corresponding sub-aperture image within the interactive circular slider.

In Figure 3, a collection of six sub-aperture images within the CSAS slider is depicted. It is worth noting that certain elements of the scene are prominently highlighted within narrow and specific ranges of aspect angles. For example, thin cables exhibit strong responses but are only clearly distinguishable over a reduced angular range, as visualized through the multi-view circular slider. In contrast, these cables are diluted and less discernible in the single-image CSAS representations. This is clearly the case for the cable connecting the central target only visible for angles close to sub-aperture in Figure 2-(b), and in a lesser extend for the cable in the upper middle sub-aperture image of Figure 3.

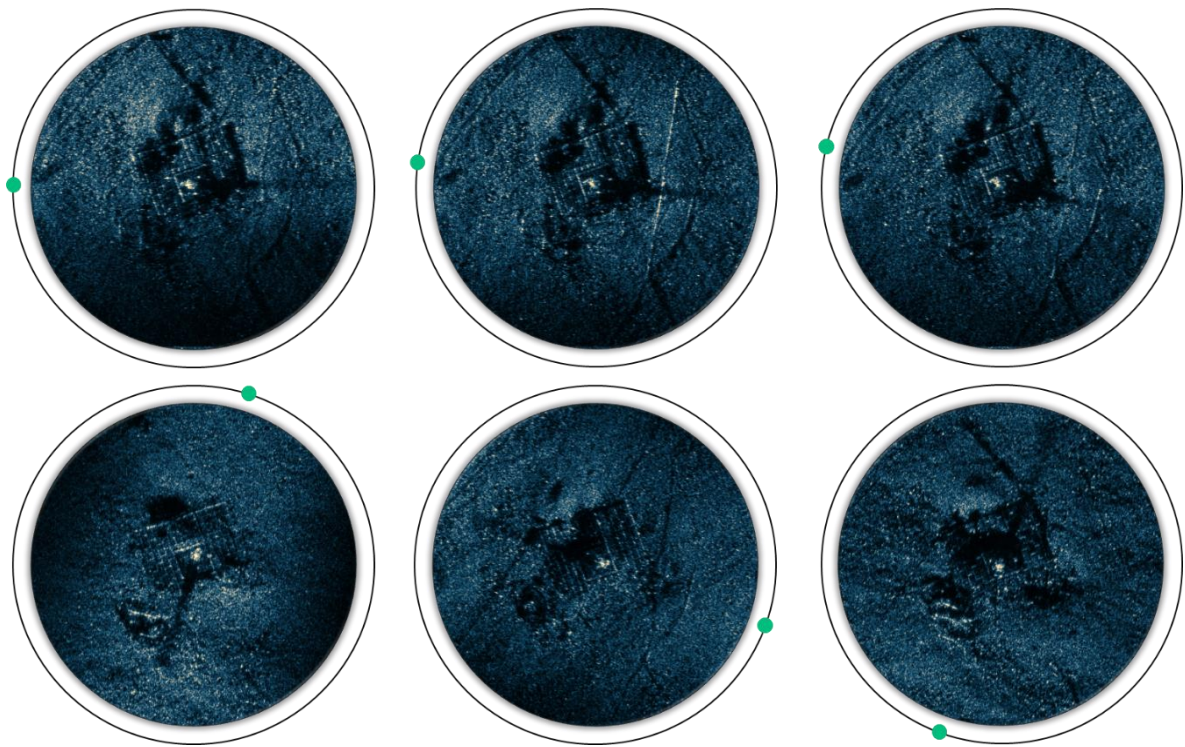

Figure 3: A collection of six sub-aperture within the CSAS slider.

Nevertheless, due to the strong parallax effect present in the sub-aperture images, the target shadow appearance tends to be degraded. To address this issue, we propose an additional multi-view slider specifically designed for shadow probing, which will be discussed in the following section.

### 3.2  Shadow enhancement

As the SAS platform moves and emits sonar pings from different positions, the incident angles of the sound waves change, leading to variations in the position of the projected shadows as shown in Figure 4. This parallax problem causes fill-in and blurring of the shadow. When the angular aperture of the antenna exceeds approximately 5 degrees, the degradation in shadow quality becomes noticeable for mine like objects. Consequently, the resulting shadow may be of poor quality, making it unsuitable for accurate target analysis.

One way to address the parallax issue is by using a smaller aperture, although this approach comes with the drawback of reduced resolution and less detailed shadows. Another approach involves compensating for the shadow's movement across the aperture. Research has demonstrated that in a linear synthetic array, parallax can be mitigated by setting the focus at the target's range [Groen et al., 2008]. This technique, known as Fixed Focus Shadow Enhancement (FFSE), can be performed either during the beamforming stage or as a post-processing step on the complex image [Sparr et al., 2007]. In the post-processing case, the following steps are involved: First, a Fast Fourier Transform (FFT) is performed along the cross-range axis of a complex image $I(x,y)$, resulting in $F(k_x, y) = FFT[I(x,y)]$. Then, for each sample $y > y_0$, where $y_0$ is the echo range, a sample-dependent filter is applied to focus at $y_0$. This filter is represented as $\tilde{F}(k_x, y) = F(k_x, y) h(k_x, y)$, where $h(k_x, y) = \exp\left(-j\Delta y \sqrt{4k_0^2 - k_x^2}\right)$, $\Delta y = y - y_0$, and $k_0 = 2\pi f_c/c$ is the wavenumber at carrier frequency. Finally, an inverse Fourier transform is performed to obtain the shadow-enhanced image $\tilde{I}(x,y) = FFT^{-1}[\tilde{F}(k_x, y)]$. A simplified expression $h(k_x, y) = \exp(-j\Delta y k_x^2 / 4k_0)$ of the filter can be also be used if the aperture is small enough.

Proceedings of the Institute of AcousticsProceedings of the Institute of Acoustics

Essentially, the operation compensates for the shadow shift $\delta x$ in the horizontal direction, as a function of the viewing angle $\theta$ and of the distance $\Delta y$ between the object and the echo. For a linear trajectory, the shadow shift is given by $\delta x = \Delta y \times \tan(\theta)$. For a circular trajectory with a radius close to the target distance the shadow horizontal shift is approximately given by $\delta x \approx \Delta y \times \sin(\theta)$ and there is also a transversal shift approximately given by $\delta y \approx \Delta y \times (1 - \cos(\theta))$. Assuming that the angle θ is small enough, which is true for limited sub-apertures, the formulas used for the linear geometry can be applied for CSAS sub-apertures as well.

The shadow migration approach, whether it follows a linear or circular trajectory, undoubtedly has some limitations. One important condition is that the shadow projection should maintain consistency across different orientations. This implies that the target's exposed surface, the bottom topography in the projected direction, and the sonar depth should exhibit minimal variation within the SAS aperture. For aperture smaller than about fifteen degrees and mine like objects, this doesn't seem to cause any major concerns most of the time.

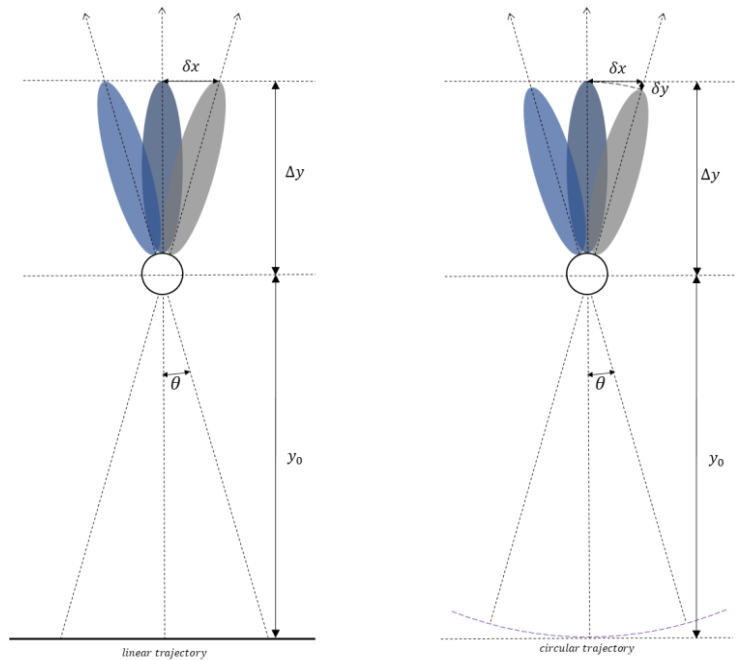

Figure 4: Illustration of the parallax effect on the shadow in synthetic aperture sonar for a linear and a circular trajectory (Upper view). The shadow is cast in different directions along the synthetic aperture and appears blurry in the final image.

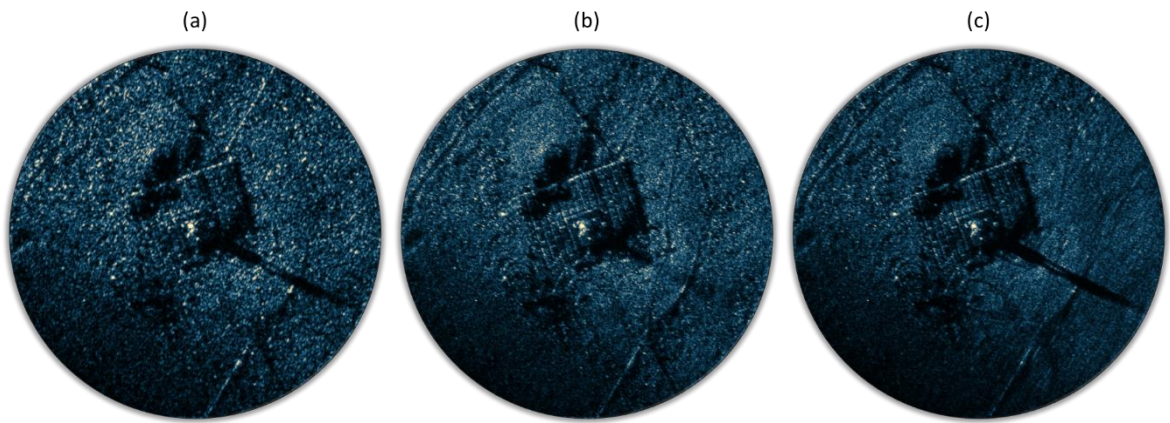

Figure 5: (a) Reduced sub-aperture image (width of 4 degrees) (b) Sub-aperture image (width of 12 degrees) (c) Shadow enhanced sub-aperture image (width of 12 degrees).

We have applied post-processing shadow enhancement as presented previously for all the sub-aperture images generated from the SAMDIS circular data. It should be noted that a preliminary rotation is necessary to perform the 1D FFT, and subsequently, an inverse rotation must be performed after the post-processing step. Figure 5-(c) provides an example, while Figure 5-(b) presents the original 12 degrees sub-aperture image for comparison. Additionally, Figure 5-(a) displays a reduced 4 degrees sub-aperture image.

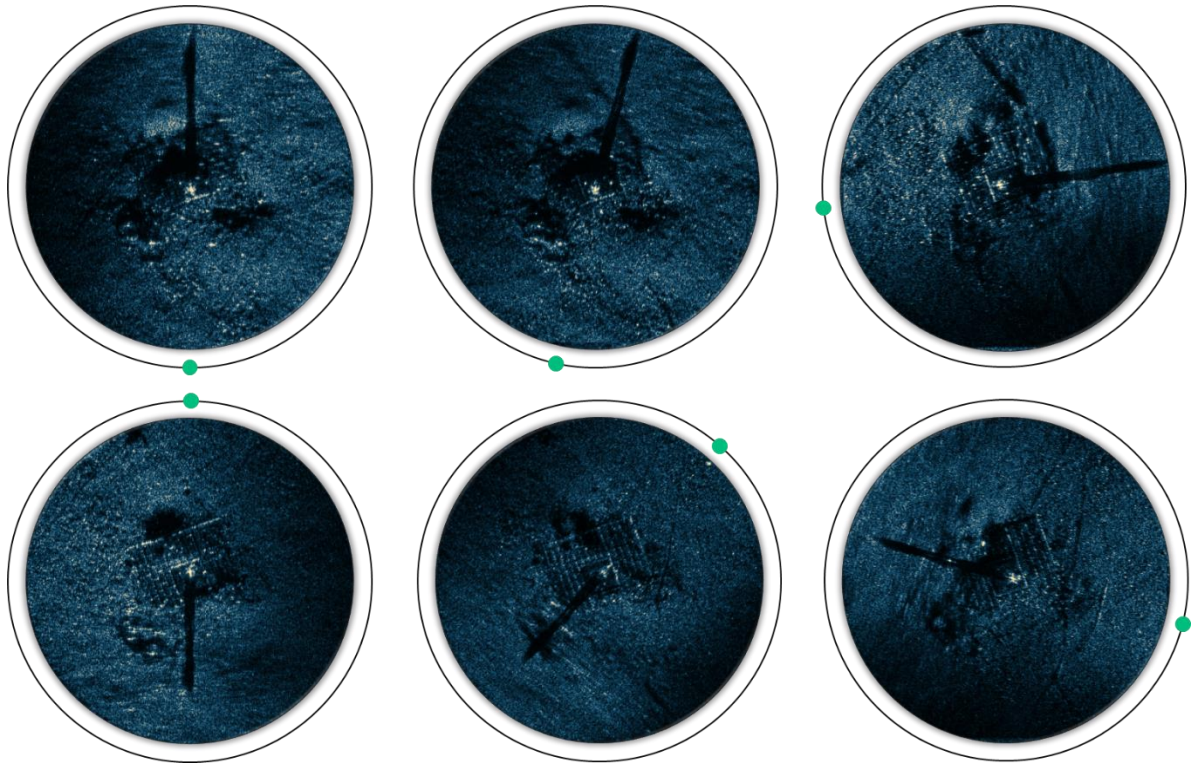

Figure 6: A collection of six shadow enhanced sub-aperture within the CSAS slider.

The resulting images has been incorporated into a circular slider dedicated for shadow analysis. A collection of six shadow enhanced sub-aperture images within the CSAS slider is depicted in Figure 6. The shape of the target can be inferred by visualizing the different shadow enhances sub-apertures with the CSAS slider. This motivate the use of 3D reconstruction method based on the shadows of the target.

## 4    3D RECONSTRUCTION

Previous research has investigated the idea of reconstructing the three-dimensional shape of an object by analysing the shadows it casts from different angles [Malkasse, 2000 ; Florin et al., 2004]. The approach becomes even more advantageous when we have a complete 360° view of the scene, as is the case with circular SAS. In the section, we explore 3D target reconstruction with space-carving methods on our shadow enhanced CSAS data.

### 4.1    Space-carving

The space-carving method is a technique used in 3D reconstruction to create a three-dimensional model of an object or scene. It involves combining multiple 2D images to generate a solid representation of the object's shape [Kutulakos et al., 2000]. In our approach, we utilize shadow segmentations of the shadow-enhanced sub-apertures as input for the space-caving algorithm. To begin, we create a 3D voxel grid that represents the volume of the target. This voxel grid is then projected onto the seafloor from an initial sub-aperture viewpoint, and compared to the segmented target shadow. If a voxel falls outside the silhouette boundaries, it is regarded as not belonging to the object and is eliminated from the grid. As we progress, more viewpoints are taken into account, and the voxel grid is further refined. The occupancy of each voxel is updated based on the consistency of projected shadows across multiple sub-apertures. By repeating this process and accumulating information from multiple viewpoints, the voxel grid gradually converges to a representation of the 3D object. Finally, the reconstructed object can be extracted by considering the occupied voxels and employing surface extraction algorithms.

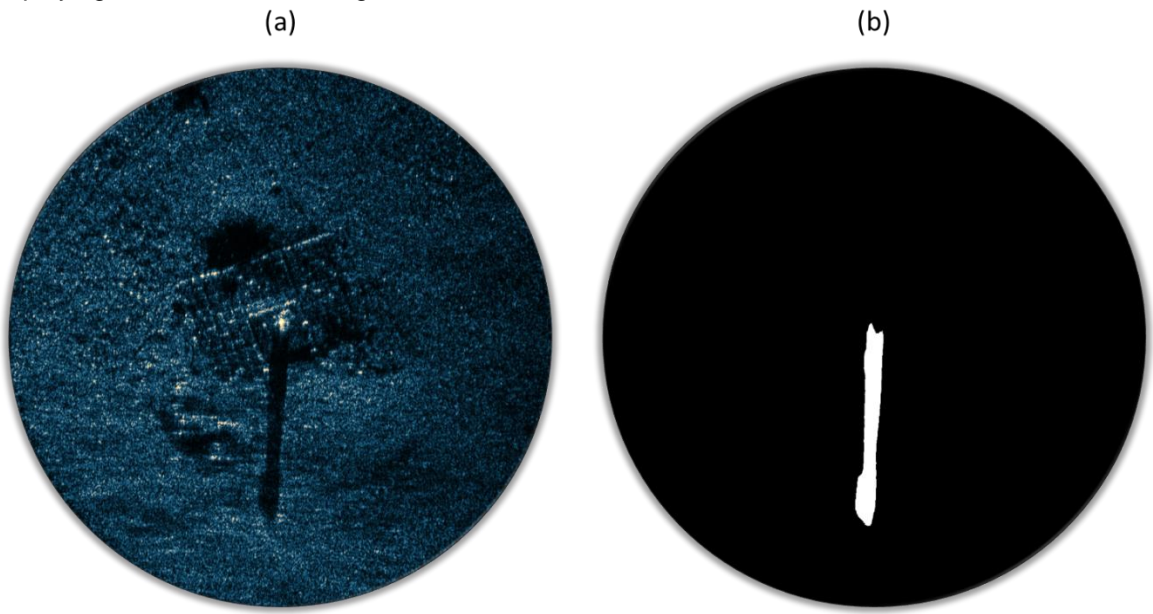

Figure 7: (a) An example of shadow enhanced sub-aperture, and (b) its manually segmented shadow.

This approach requires to extract shadows from shadow-enhanced sub-apertures, this can be done manually or automatically with segmentations algorithms. Here, for illustrative purposes, we perform manual segmentation. An example is given in Figure 7. Final result is presented in part 4.3.

### 4.2   Soft space-carving

The previous space-carving method has two significant drawbacks. Firstly, the method necessitates the segmentation of the target shadow in multiple shadow-enhanced images. Secondly, the carving operation is binary, meaning that any mistake in the shadow segmentation process results in permanent removal of voxels, thereby reducing the method's robustness. To address these limitations, we present a soft-carving approach that eliminates the need for shadow segmentation. Instead, it relies on pixel value to rate voxel likelihood.

The likelihood that a voxel lie outside the object is evaluated as follows:

$$l_{out} = \left(\frac{p_r}{p_r + p_s}\right)^2,$$

with

$$p_r = \frac{i(x,y)}{\sigma_r^2} \exp\left(-\frac{i(x,y)^2}{2\sigma_r^2}\right),$$
$$p_s = \frac{i(x,y)}{\sigma_s^2} \exp\left(-\frac{i(x,y)^2}{2\sigma_s^2}\right),$$

where $i(x,y)$ is the value of the image at voxel projected location $(x,y)$, and $\sigma_r$ and $\sigma_s$ are respectively estimates of reverberation level mode and of shadow level mode assuming Rayleigh distributions. While the validity of this assumption may not be fully confirmed, particularly in the case of shadow-enhanced images, it still proves valuable for the application. The values of $\sigma_r$ and $\sigma_s$ may be estimated on the image. Alternatively, only the reverberation mode can be estimated, and the shadow mode can be deduced from a contrast to reverberation hypothesis.

The likelihood is averaged over all the images used during the space carving procedure, and a threshold is applied at the end to get the estimated target volume. The threshold could be automatically selected, for example, based on the voxel likelihood distribution. However, for this demonstration, it has been manually set. Final result is presented in part 4.3.

## 4.3   Results

The result of the two previously described space-carving methods applied to the shadow enhanced collection of sub-aperture images is displayed on Figure 8-(a) for space carving on segmented shadows and Figure 8-(b) for soft space carving. The obtained results can be compared to the theoretical shape of the target shown in Figure 1-(a). In both cases, the volume reproduction of the target resembles the theoretical one. The dimensions are consistent too. There is no clear answer about which representation is superior, as it likely depends on the personal preferences of the operator using it. However, it is important to acknowledge that the space-carving technique applied to segmented shadows fails to capture the presence of lacunarity within the center of the target. This limitation arises from the fact that the segmented silhouettes are constructed using closed contours. On the other hand, the soft space carving is more noisy but captures the emptiness at the center of the target. It should be noted that the arms in between the two parts of the target are slightly more visible if the soft carving threshold is increased but at the cost of a more noisy representation. Both representations could be used interactively by an operator to gain insights into the shape of the target. With soft space carving, the operator could also experiment with different threshold values, providing an additional analysis capability.

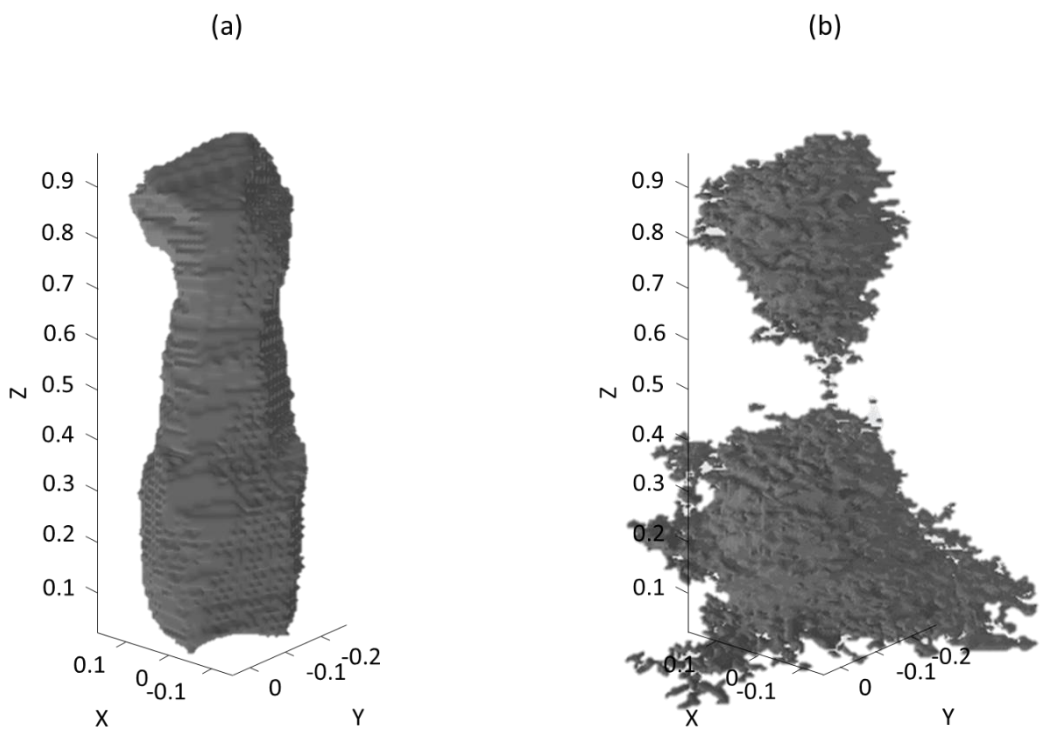

Figure 8: Space carving results on SAMDIS CSAS data, (a) space carving on segmented shadows and (b) soft space carving.

## 5    CONCLUSION

In this paper we explored an approach to enhance target analysis and achieve 3D reconstruction by retrieving shadow information from CSAS data. We employed sub-aperture filtering to capture a series of images from different viewpoints along a circular trajectory. Shadow enhancement algorithms were then applied to obtain clear and well-defined shadows. Additionally, an interactive interface was proposed to enable human operators to visualize these shadows along the circular trajectory. Finally, space-carving reconstruction methods were utilized to infer the 3D shape of the object based on the shadows.

The results highlight the potential benefit of using shadows in circular SAS for improving target analysis and 3D reconstruction. By leveraging shadow information in the analysis of CSAS data, a more comprehensive understanding of the target under observation can be obtained. Shadows can provide valuable insights into the shape and dimensions of objects, enabling more accurate target classification, identification, and interpretation. Furthermore, integrating shadow information enables the reconstruction of the target's geometry and spatial layout.

In future research, it is worth exploring more advanced techniques for 3D reconstruction from shadows, which could potentially enhance the accuracy and fidelity of the reconstructed models.

## 6    ACKNOWLEDGMENT

The authors wish to acknowledge DGA/UM-NAV, French Navy and DGA TN for their support and authorization to publish this paper.